\documentclass{article} 
\usepackage{color}
\usepackage{graphicx}
\usepackage{epstopdf}
\newtheorem{notation}{Notation}
\newtheorem{proposition}{Proposition}
\newtheorem{example}{Example}
\newtheorem{definition}{Definition}
\newtheorem{pruningrule}{Pruning Rule}
\newtheorem{feasibilityrule}{Feasibility Rule}
\begin{document}

\title{\large\bf The Soft Cumulative Constraint}


\author{ Thierry Petit        \and    Emmanuel Poder 
\and 
{\small  \texttt{~Thierry.Petit@emn.fr}~~~~\texttt{Emmanuel.Poder@laposte.net}}
\and
{\footnotesize Research report TR09/06/info,  {\'E}cole des Mines de Nantes, LINA UMR CNRS 6241.}
} 

%{$^1$  {\'E}cole des Mines de Nantes, LINA UMR CNRS 6241\\
%               4 Rue Alfred Kastler, BP 20722, 44307 Nantes Cedex 3, FRANCE\\
%              Tel.: +33 2 51 85 82 08\\\\
% %             Fax: +123-45-678910\\
%              \email{Thierry.Petit@emn.fr}           %  \\
%%             \emph{Present address:} of F. Author  %  if needed
%           E. Poder  \at
%              12 rue du Nil, 44470 Carquefou, FRANCE\\
%               \email{Emmanuel.Poder@laposte.net}  
%}

\date{}

\maketitle

\begin{abstract}
This research report presents an extension of {\em Cumulative} of Choco 
constraint solver, which is useful to  encode 
over-constrained cumulative problems. This new global constraint uses sweep 
and task interval violation-based algorithms.
\end{abstract}

\section{Introduction}
The \texttt{Cumulative} global constraint  provides  
a pruning algorithm which takes account of all 
activities at the same time, which has been proved 
to be much more efficient than CP approaches 
considering a conjunction of more primitive 
constraints. This representation as a global constraint 
has been widely studied in the
 literature and integrated into many constraint systems~\cite{aggbel93,caslab96,carottcar97,belcar02,mervan08}. 
\begin{definition}
 \label{def:activities}
Let $A=\{a_1, \ldots, a_n \}$ denote a set of $n$ non-preemptive activities. For each $a \in A$,
\begin{itemize}
\item $start[a]$  is 
the variable representing its starting point in time.
\item $dur[a] $ is the variable representing its duration. 
\item $end[a] \in end$ is the variable representing its ending point in time.
\item $res[a]$ is the variable representing the discrete amount of resource consumed by $a$, also denoted the height of $a$. 
\end{itemize}
\end{definition}
%$min(D(res[a]))$
\begin{definition} \label{def:cumulative}
Consider one resource with a limit of capacity $max\_capa$ and a set $A$ of $n$ activities. 
At each point in time $t$, the cumulated height $h_t$ of the set of
activities overlapping $t$ is $h_t = \sum_{a\in A,  start[a] \leq t < end[a]} {min(D(res[a]))}$. 

%, each one consuming a  given amount of resource. 
The \texttt{Cumulative} global 
constraint~\cite{aggbel93}  enforces that:
\begin{itemize}
\item C1: For each activity $a \in A, start[a] + dur[a] = end[a]$.
\item C2: At each point in time $t$, $h_t \leq max\_capa$. 
\end{itemize}
\end{definition}

In this research report, we deal with cumulative over-constrained 
problems that may require to be relaxed w.r.t. the resource capacity 
at some points in time, to obtain solutions. This motivates the definition 
of a new global constraint, dedicated to over-constrained instances of problems. 
\section{Cumulative Constraint with Over-Loads}  \label{sec:softCumulative} 
\begin{sloppypar}
This section presents the 
\texttt{SoftCumulativeSum} constraint,
useful to express our case-study and deal with significant instances. 
%%%%%%%%%%%%%%%%%%%%%%%%%%%%%%%%%%%%%%%%%%%%%
%\begin{notation}
$D(x)=\{min(D(x)), \ldots, max(D(x))\}$ denotes the domain of variable $x$.
%\end{notation}
\end{sloppypar}
%%%%%%%%%%%%%%%%%%%%%%%%%%%%%%%%%%%%%%%%%%%%%
%\begin{definition}\footnote{Pruning techniques presented 
%in section \ref{sec:softCumulative} can be used with activities having a duration 
%and a resource consumption which are not fixed. In this case, $min(D(dur[a]))$ and $min(D(res[a]))$ are 
%the minimum values of the corresponding domains of variables.} 
% \label{def:activities}
%Let $A$ denote a set of $n$ activities. For each $a \in A$,
%\begin{itemize}
%\item $min(D(dur[a]))$ is the value of its duration.
%\item $min(D(res[a])) \in res$ is the amount of resource
%consumed by $a$.
%\item $start[a] \in start$ is 
%the variable representing its starting point in time.
%\item $end[a] \in end$ is the variable representing its ending point
%in time (it is not mandatory to create it explicitly if durations are fixed).
%\end{itemize}
%\end{definition}
%%%%%%%%%%%%%%%%%%%%%%%%%%%%%%%%%%%%%%%
\subsection{Pruning Independent from Relaxation}
\begin{sloppypar}
The \texttt{SoftCumulativeSum} constraint that will be presented in section~\ref{subsec:relaxation}  implicitly defines a 
\texttt{Cumulative} constraint with a capacity equal to 
$max\_capa$. To prune variables in $start$,   
%from
%a partially computed schedule and its resource consumption,
several existing algorithms for \texttt{Cumulative}
 can be used. We recall the two filtering techniques 
currently implemented in the  \texttt{Cumulative} of Choco 
solver \cite{cho07}, which we extend 
to handle violations in section \ref{subsec:relaxation}. 
\end{sloppypar}
%%%%%%%%%%%%%%%%%%%%%%%%%%%%%%%%
\subsubsection{Sweep Algorithm  \cite{belcar02}}  
The goal is to reduce domains of $start$ variables 
according to the cumulated profile, which is built from compulsory parts of activities, in order not to exceed $max\_capa$. This is done in Choco using a sweep algorithm \cite{belcar02}. The \emph{Compulsory Part} \cite{lar82} of an activity is the minimum resource that will be consumed by this activity, whatever the final value of its start. It is equal to the intersection of the two particular schedules that are the activity scheduled at its latest start and the activity scheduled at its earliest start. As domains of variables get more and more restricted,  the compulsory part increases, until it becomes  the fixed activity. 

%%\vspace{-7mm}
\begin{sloppypar}
\begin{definition} \label{def:compulsory_part}
The \emph{Compulsory Part} $CP(a)$ of an activity $a$ is not empty if and only if $max(D(start[a])) < min(D(end[a]))$. If so, its height is equal to $min(D(res[a]))$ on  $[max(D(start[a])),min(D(end[a]))[$ and null elsewhere.
\end{definition}
\end{sloppypar}
%%%%
 \begin{figure}[h] 
% \vspace{-8 cm}
 \includegraphics{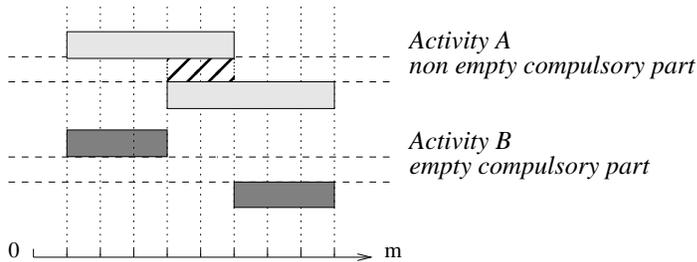}
 %\vspace{-8.5cm}
  \caption{Compulsory parts.}
   \label{fig:compulsory}
 \end{figure}
 %%%%
\begin{example}
Let $a$ be an activity such that $start[a] \in [1,4]$, with a duration fixed to $5$, 
and $b$ be an activity such that $start[b] \in [1,6]$, with a duration fixed to $3$. 
As depicted by Figure \ref{fig:compulsory}, activity $a$ has a non empty compulsory 
part, while activity $b$ has an empty compulsory part. 
\end{example}
%%%%
\begin{sloppypar}
\begin{definition} \label{def:minimum_cumulated_profile}
The \emph{Cumulated Profile} $CumP$ is the minimum cumulated resource consumption, over time, of all the activities.  For a given $t$, the height of $CumP$ at time $t$ is equal to $$\sum_{a \in A / t \in [max(D(start[a])),min(D(end[a]))[} {D(min(res[a]))}$$ That is, the sum of the contributions of all compulsory parts that overlap t.
\end{definition}
\end{sloppypar}

Next figure shows an example of a cumulative profile $CumP$ where at each point in time $t$ the height of $CumP$ at $t$ does not exceed $max\_capa$. 

\begin{figure}[h]
 %%\vspace{-8.5cm}
  \includegraphics[width=2.5in]{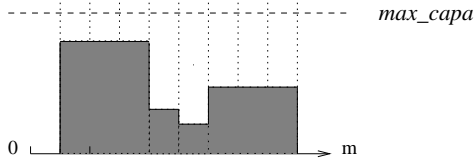}
  %%\vspace{-7.9cm}
   \caption{Cumulated profile.}
   \label{fig:profile}
 \end{figure}

%%%\vspace{-6mm}
%\noindent \textbf{Cmulated Profile}
\paragraph{Algorithm}
The sweep algorithm moves a vertical line $\Delta$ on 
the time axis from one event to the next event. In one sweep, it builds the cumulated profile 
incrementally and prunes activities according to this profile, in order not to 
exceed $max\_capa$.  
An \emph{event}  corresponds either to the start or the end of a compulsory part,  
or to the release date of an activity. All events are initially computed and sorted  in increasing 
order according to their date. Position of $\Delta$ is $\delta$. At each step of the algorithm, a 
list $ActToPrune$ contains the activities to prune. 

\begin{itemize}
\item Compulsory part events 
%that come from the compulsory parts 
are used for building $CumP$. 
All events at date $\delta$ are used 
to update the height $sum_h$ of the current rectangle in $CumP$, by 
adding the height if it is the start of a compulsory part or 
removing the height otherwise. 
%  if it is the end. 
The first %compulsory part 
event with %having 
a date strictly greater than  $\delta$ gives the end $\delta'$ of the current rectangle in $CumP$, 
finally denoted by  $\langle[\delta,\delta'[, sum_h\rangle$. 
\item Events corresponding to release dates $d$ such that $\delta \leq d <\delta'$  add some new activities to prune, according to $\langle[\delta,\delta'[, sum_h\rangle$ and $max\_capa$ (those activities which overlap $\langle[\delta,\delta'[, sum_h\rangle$). They are added to the list $ActToPrune$. 
\end{itemize}

For each $a \in ActToPrune$  that has no compulsory part registered in the rectangle 
$\langle[\delta,\delta'[, sum_h\rangle$, if its height is greater than $max\_capa - sum_h$ then the algorithm prunes  its start times so this activity doesn't overlap the current rectangle of $CumP$. Then, if the due date of $a$ is less than or equal to $\delta'$ then $a$ is removed from $ActToPrune$. After pruning activities, $\delta$  
is updated to $\delta'$. 
\begin{sloppypar}
%%%%%%%%%%%
\begin{pruningrule} \label{pruning:1} 
Let  $a \in ActToPrune$,  which has no compulsory part recorded within the  rectangle $\langle[\delta,\delta'[, sum_h\rangle$.  If $sum_h + min(D(res[a])) > max\_capa$ then $]\delta - min(D(dur[a])), \delta'[$ can be removed from $D(start[a])$. 
\end{pruningrule}
%%%%%%%%%%%
\end{sloppypar}
Time complexity of the sweep technique is $O(n*log(n))$. Please refer to the paper for more details w.r.t. this algorithm \cite{belcar02}. 
%%%%%%%%%%%%%%%%%%%%%%%%%
\subsubsection{Energy reasoning on Task Intervals \cite{lopersesq92,caslab96,baplepnuj99}}\label{subsub:teskenergy}
\begin{sloppypar}
The principle of the energy reasoning is to compare the resource  
necessarily required by a set of activities within a given interval of points in time with 
the available resource within this interval. Relevant intervals are obtained from 
starts and ends of activities (``task intervals''). This section presents 
the rules implemented in Choco.
\begin{notation}
Given $a_i$ and $a_j$ two activities (possibly the same) s.t. 
 $min(D(start[a_i])) < max(D(end[a_j]))$, we denote by: 
 \begin{itemize} 
 \item $I_{(a_i, a_j)}$ the interval $[min(D(start[a_i])), max(D(end[a_j]))[$. 
 \item $S_{(a_i, a_j)}$ the set of activities whose time-windows intersect $I_{(a_i, a_j)}$ 
and such that their earliest start is in $I_{(a_i, a_j)}$, that is, $S_{(a_i, a_j)}$ $=$ $\{ a \in A$ s.t.  $min(D(start[a_i])) \leq min(D(start[a])) < max(D(end[a_j]))\}$.
\item {\sc Area}$_{(a_i, a_j)}$ the number of free resource units in $I_{(a_i, a_j)}$, that is, 
{\sc Area}$_{(a_i, a_j)}=(max(D(end[a_j])) - min(D(start[a_i]))) * max\_capa$.
 \end{itemize}

\end{notation}
\begin{definition} \label{def:TI} 
 A lower bound $W_{(a_i, a_j)} (a)$  of the number of resource units consumed by $a \in S_{(a_i, a_j)}$ on 
$I_{(a_i, a_j)}$ is $$W_{(a_i, a_j)} (a)$$ $$=$$ $$min(D(res[a]))$$  $$*$$ $$min[min(D(dur[a])), max(0, max(D(end[a_j])) - max(D(start[a])))]$$
 \end{definition}
\end{sloppypar}
\begin{sloppypar}
\begin{feasibilityrule}\label{theo:taskInterval}
If $\sum_{a \in S_{(a_i, a_j)} }   {W_{(a_i, a_j)}}(a) >$ {\sc Area}$_{(a_i, a_j)}$ then fail.
\end{feasibilityrule}
\paragraph{Algorithm} The principle is to browse, by increasing due date,  activities $a_j \in A$ 
and for a given $a_j$ to  browse, by decreasing release date,
activities $a_i \in A$ such that $\min(D(start[a_i])) < \max(D(end[a_j]))$. 
Hence,  at each new choice of $a_i$ or $a_j$ more activities are considered.
For each couple $(a_i, a_j)$, the algorithm  applies the feasibility rule \ref{theo:taskInterval}. 

Suppose the activities sorted by increasing release date \emph{i.e.}  $\min(D(start[a_1])) \leq \min(D(start[a_2])) \leq  \cdots \leq  \min(D(start[a_n]))$, then:
\begin{itemize}
\item If $ \min(D(start[a_{i-1}])) = \min(D(start[a_{i}]))$ then\\ $S_{(a_{i-1}, a_j)} =  S_{(a_i, a_j)}$ and therefore $\sum_{a \in S_{(a_{i-1}, a_j)} }   {W_{(a_{i-1}, a_j)}}(a) = \sum_{a \in S_{(a_i, a_j)} }   {W_{(a_{i-1}, a_j)}}(a)$.\footnote{By definition, $W_{(a_i, a_j)}(a)$ is independent of $a_i$ so $W_{(a_{i-1}, a_j)}(a) = W_{(a_{i}, a_j)}(a)$.}
\item Else %($\min(D(start[a_{i-1}])) < \min(D(start[a_{i}]))$) 
\\ We have $S_{(a_{i-1}, a_j)} =  S_{(a_i, a_j)} \cup \{a_k \in A, k  \leq i-1  \wedge \min(D(start[a_k])) = \min(D(start[a_{i-1}])) \}$ \emph{i.e.} we add all activities with same release date than activity $a_{i-1}$. 
Hence, $\sum_{a \in S_{(a_{i-1}, a_j)}} {W_{(a_{i-1}, a_j)}(a)} = \sum_{a \in S_{(a_i, a_j)}} {W_{(a_i, a_j)}(a)} + \sum_{ \{k  \leq i-1  \wedge \min(D(start[a_k])) = \min(D(start[a_{i-1}])) \} } {W_{(a_i,a_j)}(a)}$. 
\end{itemize}
Therefore, the complexity for handling all intervals $I_{(a_i, a_j)}$ is $O(n^2)$. 
\end{sloppypar}
%%%%%%%%%%%%%%%%%%%%%%%%%%%%%%%%%%%%%%%%%%%%%
\subsection{Pruning Related to Relaxation} \label{subsec:relaxation}
Specific constraints  on over-loads exist in real-world applications. To express them, 
it is mandatory to discretize time, while keeping a reasonable time complexity for 
pruning. %The reason why these specific
These specific constraints are externalized because 
they are ad-hoc to each application. On the 
other hand, the following constraints capture the generic core of this class of problems. 
%For sake of expressiveness, we defined the following  
%constraint hierarchy.   
\begin{itemize}
\item \texttt{SoftCumulative} extends \texttt{Cumulative} of  
Choco by maintaining over-loads variables at each point in time, and by pruning activities according to 
maximal available capacities given by upper bounds of these variables instead of 
simply considering $max\_capa$. This constraint can be used with various  
objective criteria. 
\item The  \texttt{SoftCumulativeSum} extends the  \texttt{SoftCumulative}. It is defined to deal more efficiently with a particular criterion:  minimize the sum of over-loads. 
\end{itemize}

%%%\vspace{-6mm}
%%%%%%%%%%%%%%%%%%%%%%%%%%%%%%%%%%%%%%%%%%%%%
\subsubsection{SoftCumulative Constraint}
%%%%%%%%%%
\begin{definition} \label{def:SC}
\begin{sloppypar}
Let A be a set of activities scheduled between time $0$
and $m$, each consuming a positive amount of the resource. 
\texttt{SoftCumulative} augments \texttt{Cumulative} with  
%\begin{itemize}
%\item 
a second limit of resource $ideal\_capa \leq max\_capa$, and 
%\item  
for each point in time $t < m$ an integer variable 
$costVar[t]$. 
%\end{itemize}
It enforces: 
\end{sloppypar}
\begin{itemize}
\item C1 and C2 (see \emph{Definition \ref{def:cumulative}}).
\item C3: For each point in time $t$, $costVar[t] = max(0, h_t - ideal\_capa)$
\end{itemize}
\end{definition}
%%%%%%%%%%%%%%%%%%%%%%%%%%
\begin{figure}[h]
 %%\vspace{-7.4cm}
  \includegraphics[width=4in]{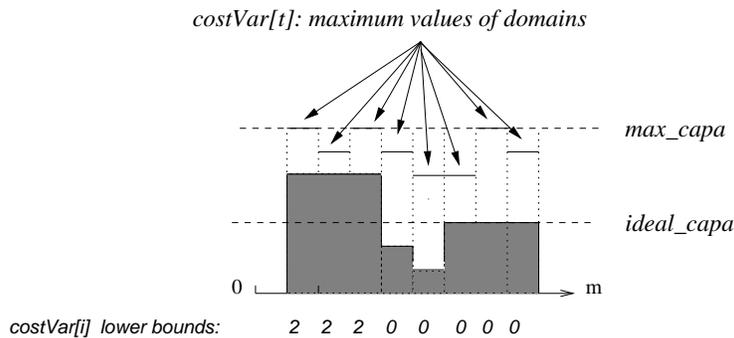}
  %%\vspace{-7cm}
   \caption{Example of a SoftCumulative constraint.}
   \label{fig:softcumulative}
 \end{figure}
%%%%%%%%%%%%%%%%%%%%%%%%%%%%%%%%%%%%%%%%%%%%%
\begin{example}
Figure \ref{fig:softcumulative} depicts an example of a cumulative profile $CumP$ where at each point in time $t$ the height of $CumP$ at $t$ does not exceed the maximum value in the 
domain of its corresponding violation variable $costVar[t]$. Time points $1$, $2$, $3$ are such that $CumP$ exceeds $ideal\_capa$ by two. 
Therefore, for each point, the minimum value of the domain of the correspondong variable in $costVar$ should be updated to value $2$.
\end{example}

\begin{sloppypar}
Next paragraph details how the classical sweep procedure can be adapted to 
the \texttt{SoftCumulative} constraint.  
\end{sloppypar}
%%\vspace{-4mm}
\paragraph{Revised Sweep pruning procedure.}  The limit of resource is not  
$max\_capa$ as in the \texttt{Cumulative} constraint. It is mandatory to take into 
account upper-bounds of variables in $costVar$. One may integrate reductions on upper bounds 
within the profile, as new fixed activities. However, our discretization of  
time can be very costly with this method : the number of events 
may grow significantly. The profile would not be computed only from 
activities but also from each point in time. 

We propose a relaxed 
version where for each rectangle we consider the maximum $costVar$ 
upper bound. This entails less pruning but the complexity is 
amortized: the number of time points checked to obtain 
maximum $costVar$ upper bounds for the whole profile is $m$, by 
exploiting the sort on events into the sweep procedure. 

Pruning rule~\ref{pruning:3} reduce domains of start variables 
from the current maximum allowed over-load in a rectangle.\footnote{The upper bound $max(D(costVar[t]))$ is the maximum value in the domain $D(costVar[t])$. Since these variables 
may be involved in several constraints, especially side constraints, the maximum value of a domain can be reduced 
during the search process.}
 
%%%%%%%%%%%
\begin{sloppypar}
\begin{pruningrule} \label{pruning:3} 
Let  $a \in ActToPrune$,  which has no compulsory part recorded within the current rectangle. 
If $sum_h + min(D(res[a])) > ideal\_capa + max_{t  \in [\delta, \delta'[}(max(D(costVar[t])))$ then $]\delta - min(D(dur[a])), \delta'[$ can be removed from $D(start[a])$. 
\end{pruningrule}
\end{sloppypar}
{\small {\bf Proof:} For any activity which overlaps the rectangle, the maximum capacity is  upper-bounded by $ ideal\_capa +  max(D(costVar[t]), t  \in [\delta, \delta'[)$. 
\\\\
}
Time complexity is $O(n*log(n)+m)$, where $m$ is the maximum due date.  

%%%%%%%%%%
\paragraph{Revised task Intervals energy reasoning.}
This paragraph describes the extension of  the principle of section~\ref{subsub:teskenergy} to 
the  \texttt{SoftCumulative} global constraint. 

%The available number of units of resource {\sc Area}$_{(a_i, a_j)}$ to consider is shrink. 
%%%%%%%%%%%
\begin{feasibilityrule}\label{theo:revistedTI1}
{\sc Area}$_{(a_i, a_j)}=$ 
$$\sum_{t \in  [min(D(start[a_i])), max(D(end[a_j]))[} ideal\_capa + max(D(costVar[t]))$$ 
If $\sum_{a \in S_{(a_i, a_j)} }   {W_{(a_i, a_j)}}(a) >$ {\sc Area}$_{(a_i, a_j)}$ then fail.
\end{feasibilityrule} 
{\small {\bf Proof:} At each time point $t$ there is $ideal\_capa + max(D(costVar[t]))$ available units. \\\\
}
Efficiency of rule \ref{theo:taskInterval} can be improved by this new computation of {\sc Area}$_{(a_i, a_j)}$ (the previous 
value was an over-estimation). Since activities $a_i$ are considered by decreasing release date, 
it is possible to compute incrementally {\sc Area}$_{(a_i, a_j)}$. Each upper bound of a variable in 
$costVar$ are considered once for each $a_j$. Time complexity is $O(n^2+n*m)$, where $m$ is the maximum due date. \\ 
%%%%%%%%%%

Next paragraph explains how minimum values of domains of variables in $costVar$ are updated during the search process.
\paragraph{Update costVar lower-bounds.}
Update of $costVar$ lower bounds can be directly performed within the sweep algorithm, 
while the profile is computed. 
%%%%%%%%%%%
\begin{sloppypar}
\begin{pruningrule} \label{update:1} 
Consider the current rectangle in the profile, $[\delta, \delta'[$. 
For each $t  \in [\delta, \delta'[$, if $sum_h - ideal\_capa > min(D(costVar[t]))$ then $[min(D(costVar[t])), sum_h - ideal\_capa[$   can be removed from $D(costVar[t])$. 
\end{pruningrule}
{\small {\bf Proof:} 
From Definitions \ref{def:compulsory_part} and \ref{def:minimum_cumulated_profile}.}\\
\end{sloppypar}
%%%%%%%%%%

Usually the constraint should not 
be associated with a search heuristic that forces to assign to a given variable in $costVar$  a value 
which is greater than the current lower bound of its domain. Indeed, such 
a search strategy would consist of imposing at this point in time a violation 
although solutions with lower over-loads at this point in time exist (or even solutions with no over-load).  
However, it is required to take into account this eventuality and 
to ensure that our constraint is valid with any search heuristic. 
If a greater value is fixed to a variable in $costVar$, until more than a very few 
number of unfixed activities exist, few deductions can be made in terms of pruning and 
they may be costly (for a quite useless feature). 
Therefore, we implemented a check procedure that 
fails when all start variables are fixed and one variable in 
$costVar$ is higher than the current profile at this point in time. 
This guarantees that ground solutions will satisfy the constraint in any case,  
with a constant time complexity. 
%%%%%%%%%%%%%%%%%%%%%%%%%%%%%%%%%%%%%%%%%%%%%
%%\vspace{-3mm}
\subsubsection{SoftCumulativeSum Constraint}\label{subsubsec:SCS}
%%%
\begin{definition} \label{def:SCS}
\begin{sloppypar}
\texttt{SoftCumulativeSum} augments \texttt{SoftCumulative}
with  an integer variable $cost$. It enforces:
\end{sloppypar}
\begin{itemize}
\item C1 and C2 (see \emph{Definition \ref{def:cumulative}}), and C3 (see \emph{Definition \ref{def:SC}}). 
\item The following constraint: $cost~~= \sum_{t \in \{0, \ldots, m-1\}} costVar[t]$
\end{itemize}
\end{definition}

\begin{sloppypar}
Pruning procedures and consistency checks of \texttt{SoftCumulative} 
remain valid for \texttt{SoftCumulativeSum}. Additionally, we aim at dealing 
with the sum constraint efficiently by exploiting the semantics.   
We compute lower bounds of the sum expressed by $cost$ variable. 
Classical back-propagation of this variable can be additionally performed as 
if the sum constraint was defined separately. 
\end{sloppypar}
\begin{example} 
The term \emph{back-propagation} is used to recall that propagation of 
events is not only performed from the decision variables 
to the objective variable but also from the objective variable to 
decision variables. For instance,  let $x_1, 
x_2$ and $x_3$ be $3$ variables with the same domain: $\forall i \in \{1,2,3\}, D(x_i) = \{1,2,3\}$. 
Let $sum$ be a variable, $D(sum) = \{3,4,\ldots,9\}$, and the following constraint $sum = \sum_{i \in \{1,2,3\}} x_i$. 
Assume that $2$ is removed from all $D(x_i)$. The usual propagation removes values $4$, $6$ and 
$8$ from $D(sum)$. Assume now that all values greater than or equal to $5$ are removed from $D(sum)$. 
Back-propagation removes value $3$ from all $D(x_i)$.
\end{example} 
%%%%%%%%%%%%%%%
\paragraph{Sweep based global lower bound.} 
Within our global constraint, a 
lower bound for the $cost$ variable is directly given 
by summing the lower bounds of all variables in $costVar$, which are  
obtained by pruning rule \ref{update:1}.  These minimum values of domains were computed from 
compulsory parts, not only from fixed activities.
$$LB_1 = \sum_{t \in \{0,\ldots,m-1\}} min(D(costVar[t]))$$
$LB_1$ can be computed with no increase in complexity 
within the sweep algorithm. 
%%%%%%%%%%%%%%%%
\paragraph{Interval based global lower bound.} 
The quantity  $\sum_{a \in S_{(a_i, a_j)} }   {W_{(a_i, a_j)}}(a)$ used in feasibility 
rule \ref{theo:taskInterval} provides the required energy for activities in the interval $I_{(a_i, a_j)}$. 
This quantity may exceed the number of time points in $I_{(a_i, a_j)}$ 
multiplied by $ideal\_capa$. We can 
improve $LB_1$, provided we remove from the computation over-loads yet taken 
into account in the $cost$ variable.  
In our implementation, we first update variables in $costVar$ 
 (by rule \ref{update:1}), and compute $LB_1$ to update $cost$. 
In this way, no additional incremental data structure is required. 

To obtain the new lower bound we need to 
compute lower-bounds of $cost$ which are local to each interval 
$I_{(a_i, a_j)}$.
%%%
\begin{definition}\label{def:dec}
$lb_{1(a_i,a_j)} = \sum_{t \in I_{(a_i, a_j)}} min(D(costVar[t]))$
\end{definition}
%%%

Then, next proposition is related to the free available number of resource units within a given interval. 
\begin{proposition}\label{lem:free}
The number {\sc FreeArea}$_{(a_i, a_j)}$ of free resource units in $I_{(a_i, a_j)}$ s. t.  
no violation is entailed is 
 $(max(D(end[a_j])) - min(D(start[a_i]))) * ideal\_capa$.
\end{proposition} 
{\small {\bf Proof:} 
From Definition \ref{def:SCS}.}\\
%%%

 From Definition \ref{def:dec} and Proposition 
\ref{lem:free},  {\sc FreeArea}$_{(a_i, a_j)} + lb_{1(a_i,a_j)}$ 
is the number of time units that can be used without 
creating any new over-load into the interval 
$I_{(a_i, a_j)}$ compared with over-loads yet taken into account 
in $ lb_{1(a_i,a_j)}$. 
\begin{definition}\label{def:inc}
%\begin{center}
{\sc Inc}$_{(a_i, a_j)} = \sum_{a \in S_{(a_i, a_j)} }   {W_{(a_i, a_j)}}(a) - ${\sc FreeArea}$_{(a_i, a_j)} - lb_{1(a_i,a_j)}$
%\end{center}
\end{definition}
{\sc Inc}$_{(a_i, a_j)}$ is the difference between 
the required energy and this quantity. 
Even if one variable in $costVar$ has a current lower bound 
higher that the value obtained from the profile, the increase {\sc Inc}$_{(a_i, a_j)}$ is valid (smaller, see Definition \ref{def:dec}). We are now able to improve $LB_1$.  
\begin{center}
$LB_2 = LB_1 + max_{(a_i, a_j) \in A^2} (${\sc Inc}$_{(a_i, a_j)})$ 
\end{center}
Another lower bound can be computed from a partition $P$ of  
$[0, m[$ in disjoint intervals obtained from pairs of activities $(a_i, a_j)$: 
$LB_{(P)} = LB_1 + \sum_{I_{(a_i, a_j)} \in P}$
 {\sc Inc}$_{(a_i, a_j)}$. Obviousy $LB_2 \leq LB_{(P)}$. However, time 
complexity of the algorithm deduced from rule 
\ref{theo:revistedTI1}  in the \texttt{SoftCumulative} constraint 
is $O(n^2+n*m)$. This complexity should reasonably not be increased. 
Computing $LB_2$ can be directly performed 
into this algorithm without any increase in complexity.\footnote{On the contrary, 
determining a relevant partition $P$ from the activities would force to 
use an independent algorithm, which can be costly depending 
on the partition we search for. Finally, we decided to use only $LB_2$. } 
%%%%%%%%%%%%%%%% 
\begin{sloppypar}
\begin{pruningrule}
If $LB_2 > min(D(cost))$ then $[min(D(cost)), LB_2[$ can be 
removed from $D(cost)$. 
\end{pruningrule}
{\small {\bf Proof:} 
From Definition \ref{def:SCS}, $LB_1$ is a lower bound of 
$cost$. Since intervals are disjoint, by Definition \ref{def:inc} the
quantity $LB_{(P)}$ is a lower bound of $cost$.  
$LB_2 \leq LB_{(P)}$. Therefore $LB_2$ is a lower bound 
of $cost$. The pruning rule holds. 
}
\end{sloppypar}
%%%%%%%%%%%%%%%
\paragraph{Aggregating local violations.} 
Once the profile is computed, if some activities having a null compulsory 
part cannot be scheduled 
without creating new over-loads, then $LB_1$ can be 
augmented with the sum of minimum increase of each activity. This idea is inspired 
from generic solving methods for over-constrained 
CSPs, e.g., Max-CSP \cite{larmes96}. Our experiments 
shown that there is quite often
a way to place any activity without creating a new violation. This 
entails a null lower bound. Therefore, we removed that computation from our implementation. 
We inform the reader that we described the procedure in a preliminary technical report~\cite{pet07}.
%%%%%%%%%%%%%%%%%%%%%%%%%%%%%%%%%%%%%%%%%%%%%

\subsubsection{Implementation}
Constraints were implemented to work with non fixed durations and 
resource consumptions.
\begin{table}
\begin{center}

\begin{small}
   \begin{tabular}{c|c|l|l}
   ~Instance~ &~$cost$ value~ &~\texttt{SoftCumulative}~ &~\texttt{SoftCumulativeSum}~ \\ 
                &                            & ~+ external sum~ &                                        \\\hline
  1   & 0  &  92 (0.07 s) & {\bf 92 (0.01 s)} \\
  2   & 2 &  417 (0.29 s) &  {\bf 94 (0.01 s)} \\
  3   & 10 &  $>$ 30 s &   {\bf 63 (0.01 s)} \\
  4   &  2 &  1301 (0.59 s) &  {\bf 194 (0.06 s)}  \\
  5   &  6 &  19506 (13.448 s)  & {\bf 97 (0.01 s)}  \\
  6   &  0 &  {\bf 53 (0.00 s)}  &  {\bf 53 (0.00 s)} \\ 
  7   &  10 &  $>$ 30 s  &   {\bf 90 (0.01 s)} \\
 8   &   6 &  $>$ 30 s  &   {\bf 152 (0.07 s)} \\
\end{tabular}

\end{small}
\caption{\small Number of nodes of {\bf optimum} schedules with $n=9$, $m=9$, 
durations between $1$ and $4$, resource consumption between $1$ and $3$, 
$ideal\_capa = 3$, $max\_capa = 7$.
}\label{tab:1}
\end{center}
%%\vspace{-1.2cm}
\end{table}
Table \ref{tab:1} compares 
the two constraints on small problems when the objective 
is to minimize $cost$. Results show  
the main importance of $LB_2$ when 
minimizing $cost$. 
%%%%%%%%%%%%%%%%%%%%%%%%%%%%%%%%%%%%%%%%%%%%
\section{Extension}\label{sec:extensions}
The global constraint presented in this research report can be tailored to be suited to some 
other classes of applications. If the time unit is tiny compared with the makespan, \emph{e.g.}, 
one minute in a one-year schedule, the same kind of model may be used 
by grouping time points. For example, each violation variable may correspond 
to one half-day. Imposing a side constraint between two particular 
minutes into a one-year schedule is generally not useful. For this purpose, the 
\texttt{SoftCumulative} constraint can be generalized, to be relaxed with respect 
to its number of violation cost variables. 
%%%
\subsection{RelSoftCumulative constraint}
\begin{sloppypar}
\begin{notation} To define \texttt{RelSoftCumulative} we use the following notations. 
Given a set of activities scheduled between $0$ and $m$,
 \begin{itemize}
 \item $mult \in \{1,2,...,m\}$ is a positive integer multiplier of the unit of time.
 \item Starting from $0$, the number of consecutive discrete intervals of length $mult$ that are included in the interval $[0,..m[$ is $\lceil m/mult \rceil$. 
 ${\cal J} =\{0, 1, \ldots, \lceil{m/mult}\rceil - 1 \}$ is the set of indexes of such intervals. Hence, to each $j \in {\cal J}$ corresponds the interval $[j * mult, j * mult + 1, \ldots  (j+1) * mult -1]$.
\end{itemize}
\end{notation}
\end{sloppypar}
\begin{definition} \label{def:RSC}
\begin{sloppypar}
Let $A$ be a set of activities scheduled between time $0$
and $m$, each consuming a positive amount of the resource. 
\texttt{RelSoftCumulative} augments \texttt{Cumulative} with  
\begin{itemize}
\item A second limit of resource $ideal\_capa \leq max\_capa$, 
\item The multiplier $mult \in \{1,2,\ldots,m\}$,
\item For each $j \in {\cal J}$ an integer variable 
$costVar[j]$. 
\end{itemize}
It enforces: 
\end{sloppypar}
\begin{itemize}
\item C1 and C2 (see \emph{Definition \ref{def:cumulative}}).
\item C4: For each $j \in {\cal J}$, $costVar[j] = \sum_{t=j*mult}^{(j+1)*mult-1} max(0, h_t - ideal\_capa)$
\end{itemize}
\end{definition}
%%%
\begin{example}
\begin{sloppypar}
We consider a cumulative over-constrained problem with $n$ activities scheduled 
minute by minute over one week. The makespan is $m = 2940$.  Assume that a user 
formulates a side constraint related to the  distribution of over-loads 
of resource among ranges of one hour ($mult=60$) in the schedule, for instance "no more 
than one hour violated each half-day".  The instance 
of \texttt{RelSoftCumulative}  related to this problem is defined with $\lceil2940/60\rceil$, {\em i.e.}, $49$ violation variables, 
${\cal J} = \{0, 1, \ldots, 48\}$. 
For each range indexed by $j \in {\cal J}$, the constraint C3 is: 
$costVar[j] = sum_{t=j*60}^{(j+1)*60-1} max(0, h_t - ideal\_capa)$. 
The side constraint is then simply expressed by cardinality constraints 
over each half-day, that is, each quadruplet of violation variables:   $\{costVar[0],\cdots,costVar[3]\}$, 
$\{costVar[4],\cdots,costVar[7]\}$, etc. 
 \end{sloppypar}
 \end{example}
\begin{sloppypar}
It is possible to reformulate rules of section \ref{subsec:relaxation} to make them 
suited to \texttt{RelSoftCumulative}. Firstly, rule \ref{pruning:3} can be re-written
 for the constraint \texttt{RelSoftCumulative}. \end{sloppypar} 
\begin{pruningrule} \label{rel:pruning:3} 
Let  $a \in ActToPrune$,  which has no compulsory part recorded within the current rectangle. \\
If $sum_h + min(D(res[a])) > ideal\_capa + max(max(D(costVar[j]), j / [\delta, \delta'[ \cap {\cal J}(j) \neq \emptyset))$\footnote{The range of 
index $j$ in ${\cal J}$ corresponding to time point $t$ is $\lfloor t/mult \rfloor$. Hence, the set of $j$ such that $[\delta, \delta'[ \cap {\cal J}(j) \neq \emptyset))$ 
is $\{ \lfloor \delta/mult \rfloor,\ldots, \lfloor (\delta'-1)/mult \rfloor\}$.}
 then $]\delta - min(D(dur[a])), \delta'[$ can be removed from $D(start[a])$. 
\end{pruningrule}
Similarly, rule \ref{theo:revistedTI1} is reformulated as follows:
 \begin{feasibilityrule}
 {\sc Area}$_{(a_i, a_j)}=$ 
$$\sum_{t \in  [min(D(start[a_i])), max(D(end[a_j]))[} ideal\_capa + max(D(costVar[\lfloor t/mult \rfloor]))$$ 
If $\sum_{a \in S_{(a_i, a_j)} }   {W_{(a_i, a_j)}}(a) >$ {\sc Area}$_{(a_i, a_j)}$ then fail.
\end{feasibilityrule} 
\begin{sloppypar}
To update minimum values of domains of variables in $costVar$ in  
\texttt{RelSoftCumulative}, we simply have to update for each 
violation variable, during the sweep, the current sum of over-loads 
of its time points. This may be done only by maintaining one 
value and one index, but for sake of clarity we use the following notation.
\begin{notation}
Given a set of ranges in time indexed by ${\cal J} = \{0, 1, \ldots, \lceil{m/mul}\rceil - 1 \}$, 
\emph{\texttt{costarray}} is an array of integers. All of them are initially set to $0$. They 
are one-to-one mapped 
with elements in ${\cal J}$.
\end{notation}
Next rule reformulates rule \ref{update:1} for \texttt{RelSoftCumulative}.
\end{sloppypar}
\begin{sloppypar}
\begin{pruningrule} \label{rel:update:1} 
Consider the current rectangle in the profile, $[\delta, \delta'[$. 
For each $t  \in [\delta, \delta'[$, if $sum_h - ideal\_capa > 0$ then:
\begin{enumerate}
\item \emph{\texttt{costarray}[$\lfloor t/mult \rfloor$]} $\leftarrow$ \emph{\texttt{costarray}[$\lfloor t/mult \rfloor] +  sum_h - ideal\_capa$}
\item if \emph{\texttt{costarray}[$\lfloor t/mult \rfloor$]} $> min(D(costVar[\lfloor t/mult \rfloor]))$ then:\\ The range $[min(D(costVar[\lfloor t/mult \rfloor]))$, \emph{\texttt{costarray}[$\lfloor t/mult \rfloor]$}$[$ can be removed from $D(costVar[\lfloor t/mult \rfloor])$. 
\end{enumerate}
\end{pruningrule}
\end{sloppypar}
%%%
\subsection{RelSoftCumulativeSum constraint}
\begin{definition} \label{def:RSCS}
\begin{sloppypar}
\texttt{RelSoftCumulativeSum} augments \texttt{RelSoftCumulative}
 with  an integer variable $cost$. It enforces:
\end{sloppypar}
\begin{itemize}
\item C1 and C2 (see \emph{Definition \ref{def:cumulative}}), and C4 (see \emph{Definition \ref{def:RSC}}). 
\item The following constraint: $cost~~= \sum_{j \in {\cal J}} costVar[j]$
\end{itemize}
\end{definition}
\paragraph{Sweep based global lower bound.} 
as for the \texttt{SoftCumulativeSum} constraint presented in 
section \ref{subsubsec:SCS}, a lower bound for the $cost$ 
objective variable is given 
by summing the lower bounds of all variables in $costVar$, 
without any increase in complexity in the sweep algorithm. 
$$LB_1 = \sum_{j \in {\cal J}} min(D(costVar[j]))$$
\paragraph{Interval based global lower bound.} 
The task interval energetic reasoning presented in section \ref{subsubsec:SCS} 
remains the same, except the evaluation of the quantity $lb_1$, 
which corresponds to over-loads  
expressed by variables in $costVar$ array (see definition \ref{def:dec}). 

Within the filtering algorithm of the \texttt{SoftCumulativeSum} constraint, 
$lb_1$ is the sum of $min(costVar[t])$ over all 
points int time $t$ within each considered interval, providing that, in the 
implementation, all variables in $costVar$ array are updated 
before computing $lb_1$. 
With respect to the \texttt{RelSoftCumulative} constraint, 
by definition \ref{def:RSCS}, if $mult$ is greater than or equal to two then  
it is not possible to evaluate at each point in time $t$ the 
exact over-load at $t$.
Under-estimating $lb_1$ would be false because this leads to a over-estimation 
of task interval based lower bounds (see Definition \ref{def:inc}). 

Therefore, 
we compute an over-estimation of $lb_1$, the tightest possible 
according to the definition of the constraint. 
\begin{notation} 
Let $I=[a,b]$ be an interval of points in time included in $[0,m[$, and 
the set of indexes for intervals in time ${\cal J} = \{0, 1, \ldots, \lceil{m/mul}\rceil - 1 \}$. 
For each $j \in J$, $\#(I,j)$ is the number of time points in common between 
the range indexed by $j$ and $I$: $$\#(I,j) =| [a, b] \cap {\cal J}(j)|$$  
\end{notation}

We can now reformulate Definition \ref{def:dec} for the 
\texttt{RelSoftCumulative}. The idea is to evaluate for each ${\cal J}(j)$ intersecting 
the current interval $I_{(a_i, a_j)}$, the minimum value between $min(D(costVar[j]))$ and 
the maximum possible over-load in $I \cap  {\cal J}(j)$, which is 
equal from Definition~\ref{def:RSCS} to $\#(I,j) * (max(D(costVar[j])) - ideal\_{capa})$. 
\begin{sloppypar}
\begin{definition} 
Let $I_{(a_i,a_j)}$ be a task interval, and $J = \{j \in {\cal J},  \#j(I_{(a_i,a_j)}) > 0 \}$.
$$lb_1 = \sum_{j \in J} min( min(D(costVar[j])),  \#(I,j) * (max(D(costVar[j])) - ideal\_{capa}))$$
\end{definition}
 \end{sloppypar}
%%%%%%%%%%%%%%%%%%%%%%%%%%%%%%%%%%%%%%%%%%%%
\section{Conclusion}
This report 
proposed several filtering procedures for a global \texttt{Cumulative} constraint 
which is relaxed w.r.t. to its capacity at some points in time. 
We provided the extension of our global constraint 
for the case where side constraints are related to ranges 
in time which are larger than one time unit.

\bibliographystyle{spmpsci}     
\bibliography{AOR08new}   
\end{document}